\documentclass[sigconf]{acmart}

\usepackage[utf8]{inputenc}   
\usepackage[T1]{fontenc}      

\usepackage{amsmath, amssymb} 
\usepackage{dsfont}           

\usepackage{booktabs}         
\usepackage{nicefrac}         
\usepackage{microtype}        
\usepackage{etoolbox}         
\usepackage{wrapfig}          
\usepackage{subcaption}       
\usepackage{caption}          
\usepackage{pifont}           
\usepackage{array, multirow, graphicx} 
\usepackage{subcaption}
\usepackage{algorithm}
\usepackage{algorithmic}
\usepackage{listings}         
\usepackage{url}              
\usepackage{tikz}             
\usepackage{comment}          
\usepackage{cancel}           

\AtBeginDocument{%
  }

\acmYear{2025}

\begin{document}

\title{ACM Multimedia Grand Challenge on ENT Endoscopy Analysis}

\author{Trong-Thuan Nguyen}
\orcid{0000-0001-7729-2927}
\affiliation{%
  \institution{University of Science, VNU-HCM}
  \city{Ho Chi Minh}
  \country{Vietnam}
}
\affiliation{%
  \institution{Vietnam National University}
  \city{Ho Chi Minh}
  \country{Vietnam}
}
\email{ntthuan@selab.hcmus.edu.vn}

\author{Viet-Tham Huynh}
\orcid{0000-0002-8537-1331}
\affiliation{%
  \institution{University of Science, VNU-HCM}
  \city{Ho Chi Minh}
  \country{Vietnam}
}
\affiliation{%
  \institution{Vietnam National University}
  \city{Ho Chi Minh}
  \country{Vietnam}
}
\email{hvtham@selab.hcmus.edu.vn}

\author{Thao Thi Phuong Dao}
\orcid{0000-0002-0109-1114}
\affiliation{%
  \institution{University of Science, VNU-HCM}
  \city{Ho Chi Minh}
  \country{Vietnam}
}
\affiliation{%
  \institution{Vietnam National University}
  \city{Ho Chi Minh}
  \country{Vietnam}
}
\affiliation{%
  \institution{Thong Nhat Hospital}
  \city{Ho Chi Minh}
  \country{Vietnam}
}
\email{dtpthao@selab.hcmus.edu.vn}

\author{Ha Nguyen Thi}
\orcid{0009-0006-1428-5504}
\affiliation{%
  \institution{Thong Nhat Hospital}
  \city{Ho Chi Minh}
  \country{Vietnam}
}
\email{hant3@bvtn.org.vn}

\author{Tien To Vu Thuy}
\orcid{0009-0002-8378-8827}
\affiliation{%
  \institution{Faculty of Medicine, \\ Pham Ngoc Thach University of Medicine}
  \city{Ho Chi Minh}
  \country{Vietnam}
}
\email{tientvt@pnt.edu.vn}

\author{Uyen Hanh Tran}
\orcid{0000-0002-5274-631X}
\affiliation{%
  \institution{Cho Ray Hospital}
  \city{Ho Chi Minh}
  \country{Vietnam}
}
\email{uyenent@gmail.com}

\author{Tam V. Nguyen}
\orcid{0000-0003-0236-7992}
\affiliation{%
  \institution{University of Dayton}
  \city{Ohio}
  \country{United States}
}
\email{tnguyen1@udayton.edu}

\author{Thanh Dinh Le}
\orcid{0009-0009-3153-085X}
\affiliation{%
  \institution{University of Health Sciences, VNU-HCM}
  \city{Ho Chi Minh}
  \country{Vietnam}
}
\affiliation{%
  \institution{Vietnam National University}
  \city{Ho Chi Minh}
  \country{Vietnam}
}
\affiliation{%
  \institution{Thong Nhat Hospital}
  \city{Ho Chi Minh}
  \country{Vietnam}
}
\email{ledinhthanhvmc@yahoo.com.vn}

\author{Minh-Triet Tran}
\orcid{0000-0003-3046-3041}
\affiliation{%
  \institution{University of Science, VNU-HCM}
  \city{Ho Chi Minh}
  \country{Vietnam}
}
\affiliation{%
  \institution{Vietnam National University}
  \city{Ho Chi Minh}
  \country{Vietnam}
}
\email{tmtriet@fit.hcmus.edu.vn}

\renewcommand{\shortauthors}{Trong-Thuan Nguyen et al.}

\begin{abstract}
Automated analysis of endoscopic imagery is a critical yet underdeveloped component of ENT (ear, nose, and throat) care, hindered by variability in devices and operators, subtle and localized findings, and fine-grained distinctions such as laterality and vocal-fold state. In addition to classification, clinicians require reliable retrieval of similar cases, both visually and through concise textual descriptions. These capabilities are rarely supported by existing public benchmarks. To this end, we introduce ENTRep, the ACM Multimedia 2025 Grand Challenge on ENT endoscopy analysis, which integrates fine-grained anatomical classification with image-to-image and text-to-image retrieval under bilingual (Vietnamese and English) clinical supervision. Specifically, the dataset comprises expert-annotated images, labeled for anatomical region and normal or abnormal status, and accompanied by dual-language narrative descriptions. In addition, we define three benchmark tasks, standardize the submission protocol, and evaluate performance on public and private test splits using server-side scoring. Moreover, we report results from the top-performing teams and provide an insight discussion.
\end{abstract}

%
%




\maketitle


\section{Introduction}\label{sec:intro}
Endoscopic imaging constitutes a critical modality in otolaryngology (ENT)~\cite{wu2023advances, ali2024demographic, demir2025artificial}, enabling the visual assessment of anatomical structures such as the external auditory canal and tympanic membrane, nasal cavities and septum, and the laryngeal inlet, including the vocal folds. Despite its clinical ubiquity, the automated interpretation of ENT endoscopic imagery remains a substantial challenge. Image characteristics are influenced by numerous factors, including the type of endoscope used, lighting conditions, and operator technique. Moreover, diagnostically relevant features are often subtle, spatially localized, and distributed unevenly across anatomical regions and pathological categories, resulting in a pronounced long-tailed label distribution. The need to resolve fine-grained distinctions, such as laterality (left versus right) or vocal fold positioning (open versus closed), further exacerbates the complexity, revealing failure cases inadequately captured by standard benchmarks.

Recent developments in vision–language modeling have demonstrated that aligning visual data with natural language supervision can substantially improve open-vocabulary recognition and enable more versatile retrieval systems~\cite{hu2022x, chen2024bimcv}. However, the majority of such advances have been validated on general datasets or within medical subfields characterized by greater availability of public data. ENT endoscopy, despite its clinical significance, remains underexplored in this context. In practical settings, the capacity to retrieve relevant clinical cases via either an image or text is often as diagnostically informative as assigning a fixed label, facilitating case-based reasoning, medical education, and rapid expert verification.

To address this gap, the ENTRep Challenge~\footnote{\url{https://aichallenge.hcmus.edu.vn/acm-mm-2025/entrep}} (Advancing Vision-Language AI for ENT Endoscopy Analysis), organized in conjunction with ACM Multimedia (MM) 2025, introduces a new benchmark dataset comprising ENT endoscopic images acquired at Thong Nhat Hospital (Ho Chi Minh City, Vietnam). Specifically, each image is annotated with expert-confirmed anatomical region labels and accompanied by bilingual clinical descriptions in Vietnamese and English. In addition, our grand challenge encompasses three tasks that reflect typical clinical workflows: (1) image classification, (2) image-to-image retrieval, and (3) text-to-image retrieval. 

To the best of our knowledge, ENTRep represents the first benchmark to jointly evaluate fine-grained ENT classification alongside both intra-modal (image$\rightarrow$image) and cross-modal (text$\rightarrow$image) retrieval under bilingual clinical supervision. By centering evaluation on tasks that align with real-world diagnostic practices and by providing a transparent, reproducible evaluation framework, ENTRep aims to establish a standardized basis for algorithmic benchmarking and to facilitate the translation of advances in vision–language modeling to clinically meaningful applications.

This paper formalizes the challenge tasks and evaluation metrics, details the dataset schema and usage policies, and outlines the organizational structure of the competition. In addition, we present baseline results on both public and private test splits and conclude by discussing open research challenges and future directions.
\begin{figure}[!t]
    \centering
    \includegraphics[width=\linewidth]{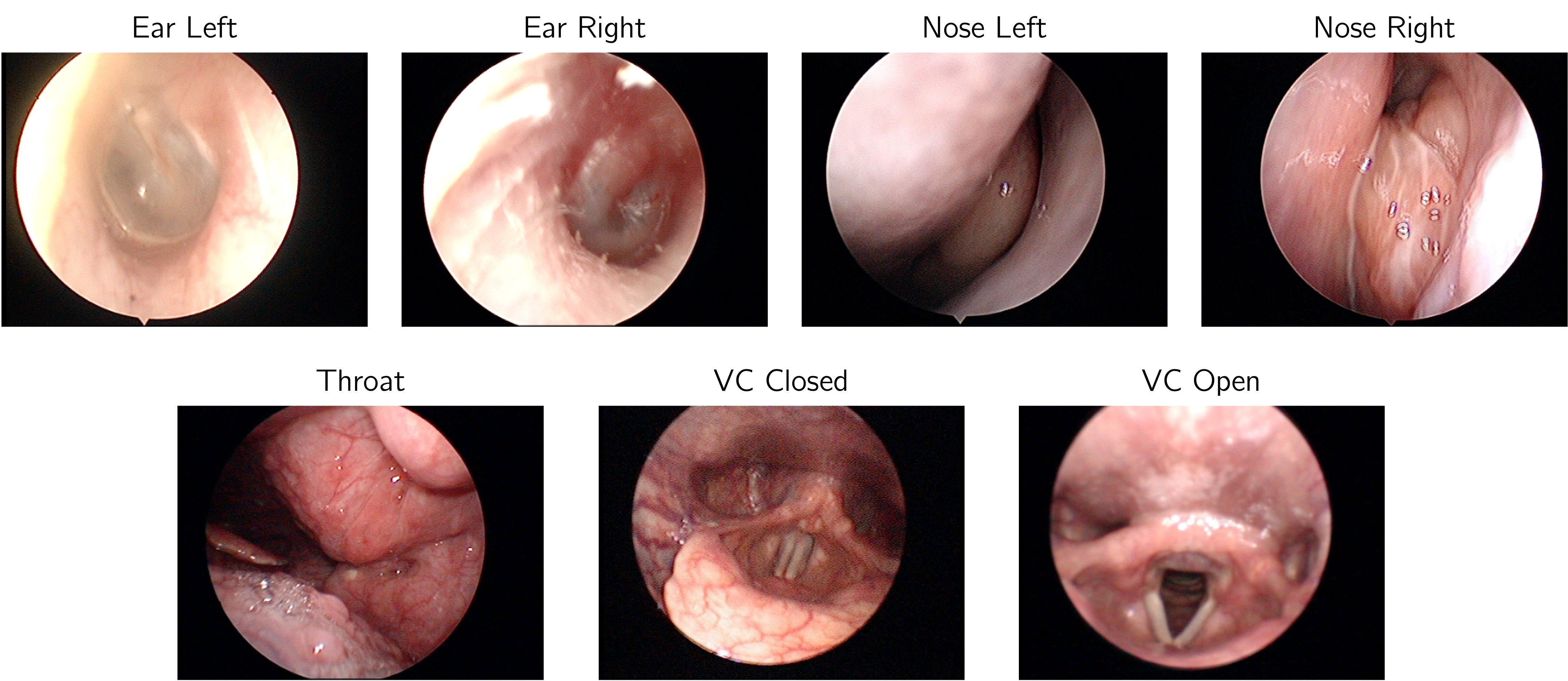}
    \caption{Task~1 (\textit{Image Classification}): \texttt{Ear Right}, \texttt{Ear Left}, \texttt{Nose Right}, \texttt{Nose Left}, \texttt{Throat}, \texttt{VC-open}, and \texttt{VC-closed}. The goal is to assign one region label to each endoscopic image.}
    \label{fig:cls}
\end{figure}

\section{Challenge Tasks}\label{sec:tasks}
We define three challenge tracks, in which participating teams may engage in any track. All submissions adhere to a unified output schema across the public test and private test phases. Evaluation is conducted server-side to ensure consistency and fairness in scoring.

\subsection{Task 1: Image Classification}
\textit{Objective.} Given a single ENT endoscopic image, predict its anatomical region. The seven region types are illustrated in Fig.~\ref{fig:cls}.

\textit{Input/Output.} The input is an RGB image; the output is a categorical label from the following categories: \texttt{Ear Right}, \texttt{Ear Left}, \texttt{Nose Right}, \texttt{Nose Left}, \texttt{Throat}, \texttt{VC-open}, and \texttt{VC-closed}.

\textit{Clinical Relevance. }Supports automated image tagging, triage, and dataset organization in clinical workflows and archival systems.

\subsection{Task 2: Image-to-Image Retrieval}
\begin{figure}[!t]
    \centering
    \includegraphics[width=\linewidth]{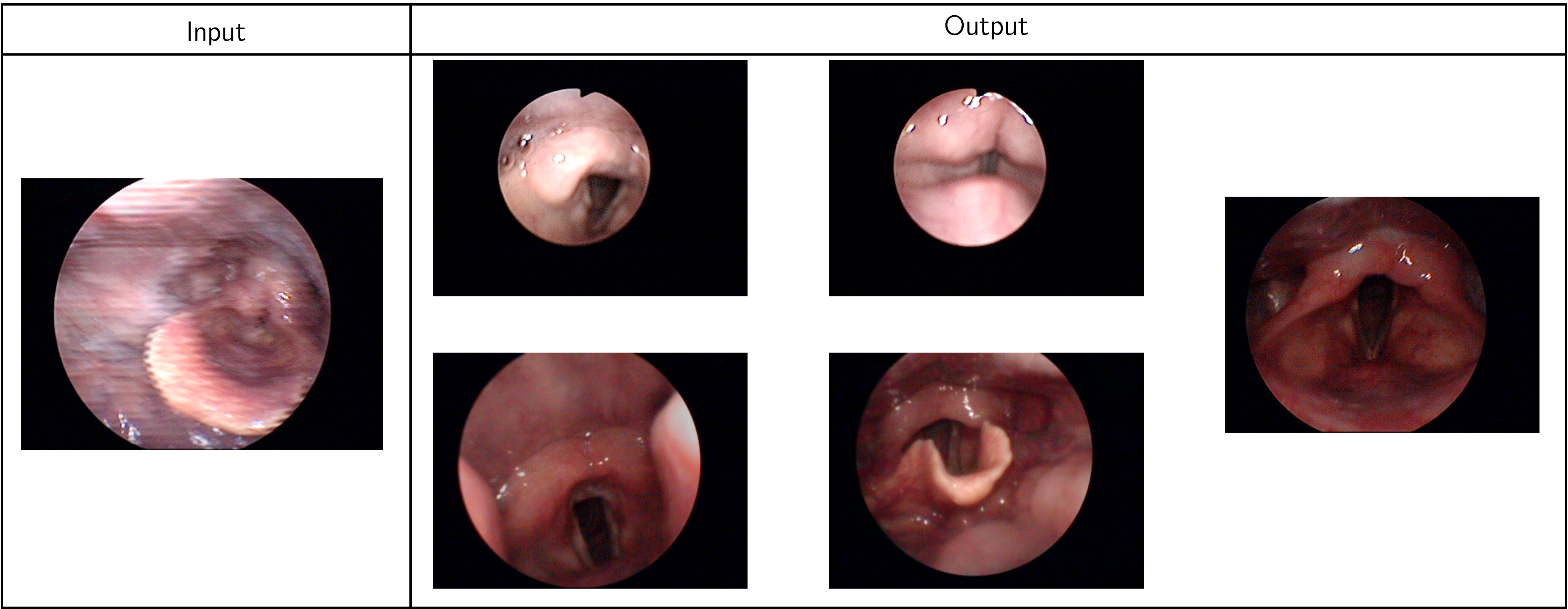}
    \caption{Task~2 (\textit{Image-to-Image Retrieval}): given an image, the system returns a ranked list of database images.}
    \label{fig:t2i}
\end{figure}

\textit{Objective.} Given a query image, retrieve a ranked list of visually or semantically similar images from the dataset (illustrated in Fig.~\ref{fig:t2i}).

\textit{Input/Output.} The input is an RGB query image; the output is a ranked list of images corresponding to the most relevant matches.

\textit{Clinical Relevance.} Enables visual case-based retrieval to support differential diagnosis, clinical training, and comparative analysis.

\subsection{Task 3: Text-to-Image Retrieval}
\begin{figure}[!t]
    \centering
    \includegraphics[width=\linewidth]{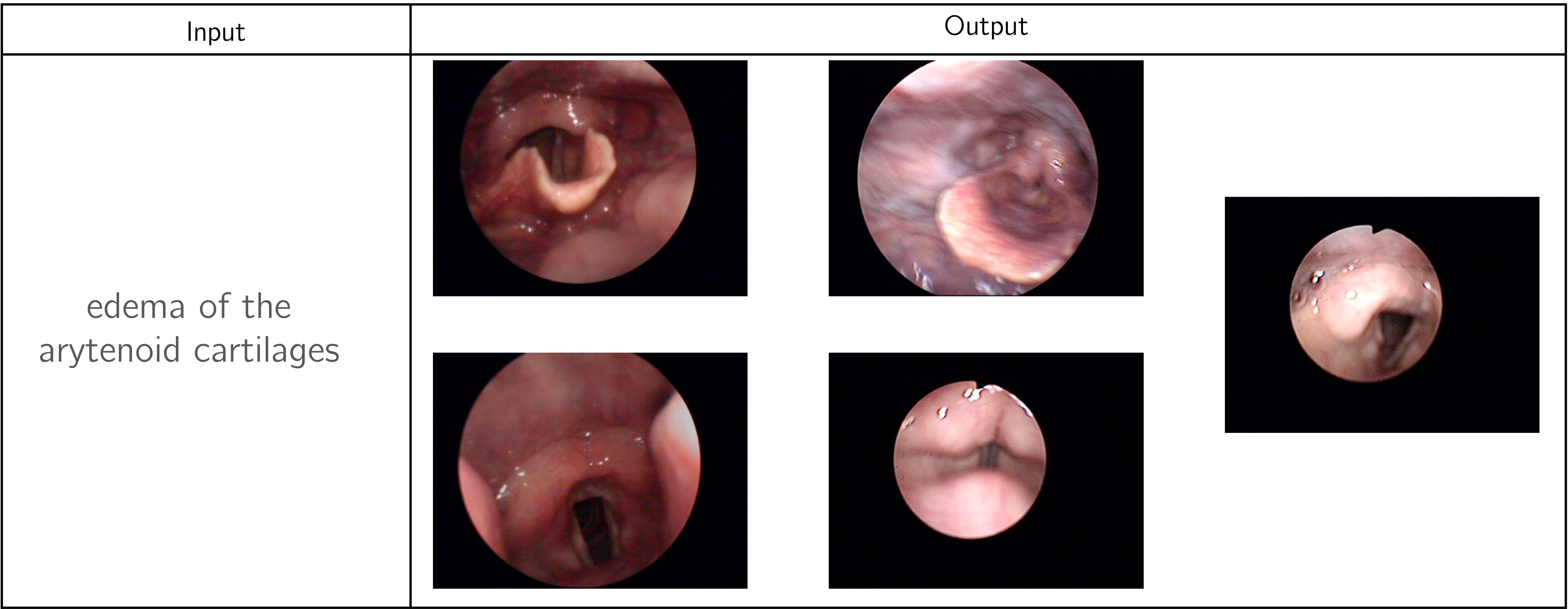}
    \caption{Task~3 (Text-to-Image Retrieval): given a clinical description, the system retrieves a ranked list of images.}
    \label{fig:i2i}
\end{figure}

\textit{Objective.} Given an English-language clinical description, retrieve the most relevant images from the dataset (illustrated in Fig.~\ref{fig:i2i}).

\textit{Input/Output.} The input is a query (textual description); the output is a ranked list of matching images from the dataset.

\textit{Relevance Criterion.} An image is considered relevant if it corresponds to the ground-truth image(s) annotated for the query.

\textit{Clinical Relevance.} Facilitates cross-modal case retrieval, allowing clinicians to find representative ENT endoscopic images using concise textual descriptions, aiding both education and diagnosis.

\section{Methods Overview}\label{sec:method}
\begin{figure*}[!t]
    \centering
    \includegraphics[width=1\linewidth]{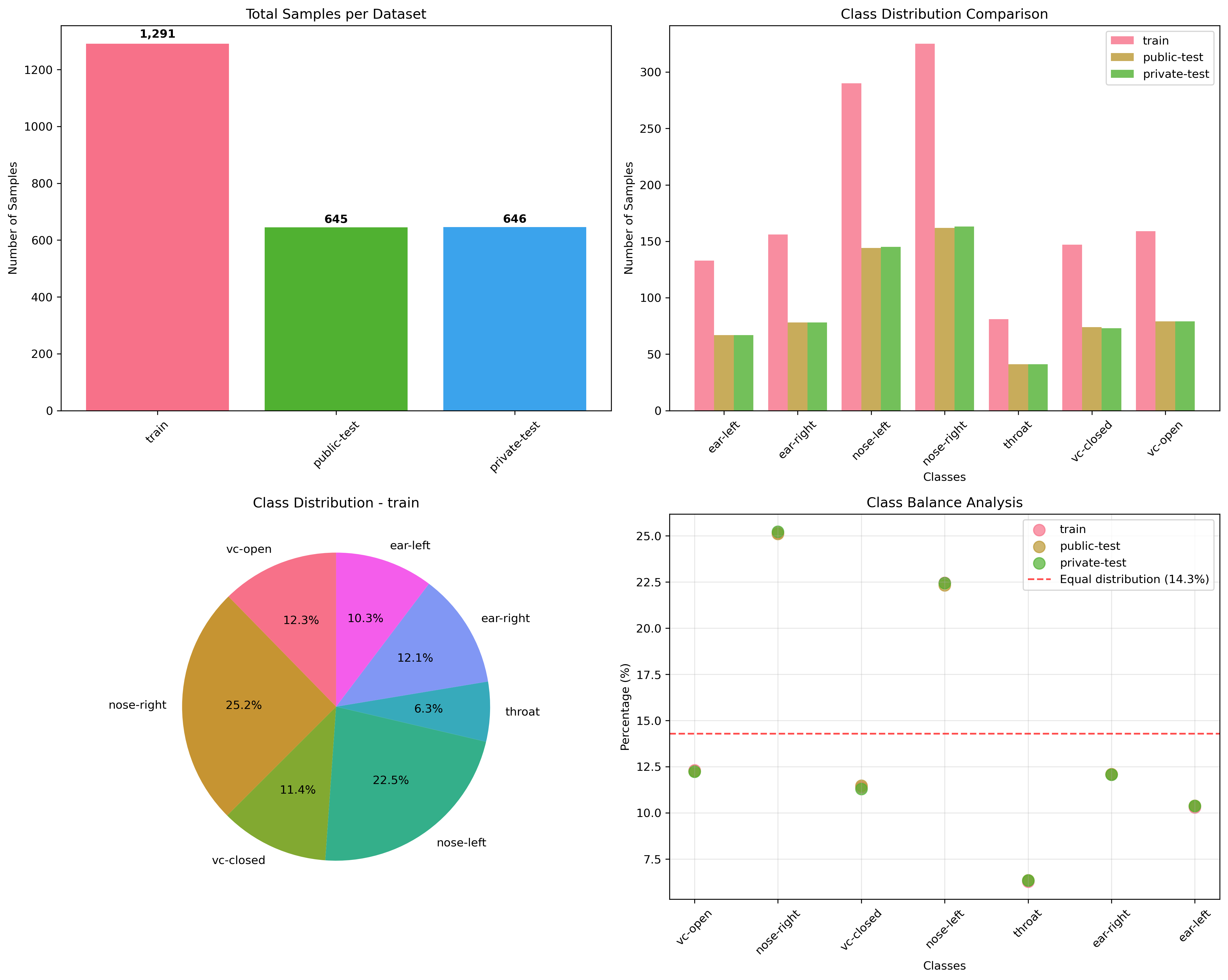}
    \caption{Overview of dataset partitioning and class distribution for the classification task.}
    \label{fig:cls_stats}
\end{figure*}

Across all three ENTRep tracks, top-performing submissions converge on a common strategy: they start with strong pre-trained encoders, adapt them cautiously to the small and imbalanced endoscopy domain, and separate recall-oriented initial stages from precision-focused refinement. Specifically, teams favor staged or parameter-efficient fine-tuning to preserve generic visual or vision–language priors while incorporating anatomy-specific signals. Data handling remains anatomy-aware: augmentations respect critical attributes such as laterality (left/right) and midline or state labels (e.g., vocal cords open/closed), and sampling strategies or loss shaping address class imbalance without compromising probability calibration. Where latency permits, teams apply modest ensembling and test-time augmentation to stabilize predictions.

For \textbf{\textit{Track 1 (Image Classification)}}, leading methods adapt modern CNN and ViT backbones using phased unfreezing and conservative learning rates. They begin with partial fine-tuning and gradually progress to full fine-tuning. Augmentations target endoscopic artifacts, such as light jitter, blur, small rotations, and denoising, and anatomical considerations, including horizontal flips with label swapping for left/right classes and flips without relabeling for throat and vocal-cord states. Several teams introduce class-preserving mixing strategies, such as intra-class mixup or mosaics, to enhance intra-class diversity. To address class imbalance, methods apply weighted sampling and focal-style loss functions with label smoothing. During inference, most approaches use light test-time augmentation and apply soft voting across cross-validation folds or diverse backbones where computationally feasible.

For \textbf{\textit{Track 2 (image-to-image retrieval)}}, high-performing teams use a two-stage design. A bi-encoder generates gallery embeddings offline and enables fast approximate nearest-neighbor search to retrieve top-$K$ candidates at query time. In the second stage, the system enhances clinical precision through reranking, either using a stronger similarity function or incorporating lightweight multimodal reasoning guided by ENT landmarks and terminology. To reconcile coarse visual similarity with fine-grained clinical semantics, teams apply position-aware fusion that adjusts the initial rank based on refined scores. Many methods also gate the gallery using a coarse anatomical prediction to reduce symmetric confusions, such as left versus right, and to mitigate lighting-induced false positives.

For \textbf{\textit{Track 3 (text-to-image and image-to-text retrieval)}}, methods map text and images into a shared embedding space using domain-aware vision–language models, then add semantic scaffolding to improve alignment. In particular, this includes concise anatomy-aligned prompts or descriptors when free-text input is sparse, bilingual querying when beneficial, and a reranking stage that grounds retrieval decisions in clinically relevant cues. As in Track 2, classifier-informed pruning and rank–score fusion are used to enhance top-rank precision while maintaining minimal loss.

Common across all tracks are gentle endoscopy-specific preprocessing methods, such as contrast normalization and vignetting or region-of-interest cropping. Specifically, most teams use stratified K-fold validation with early stopping and cosine-style learning rate scheduling, alongside a consistent focus on computational efficiency. Embeddings and indices are precomputed to ensure that inference latency is dominated by the small value of $K$ used during reranking. The degree of test-time augmentation or ensembling is carefully adjusted to meet deployment constraints. Overall, the strongest submissions combine robust pre-trained priors with anatomy-aware adaptation and modular second-stage refinement, achieving a balance of speed, stability, and clinical relevance.

\section{Dataset and Evaluation Metric}\label{sec:data}

\begin{figure}[!t]
    \centering
    \includegraphics[width=\linewidth]{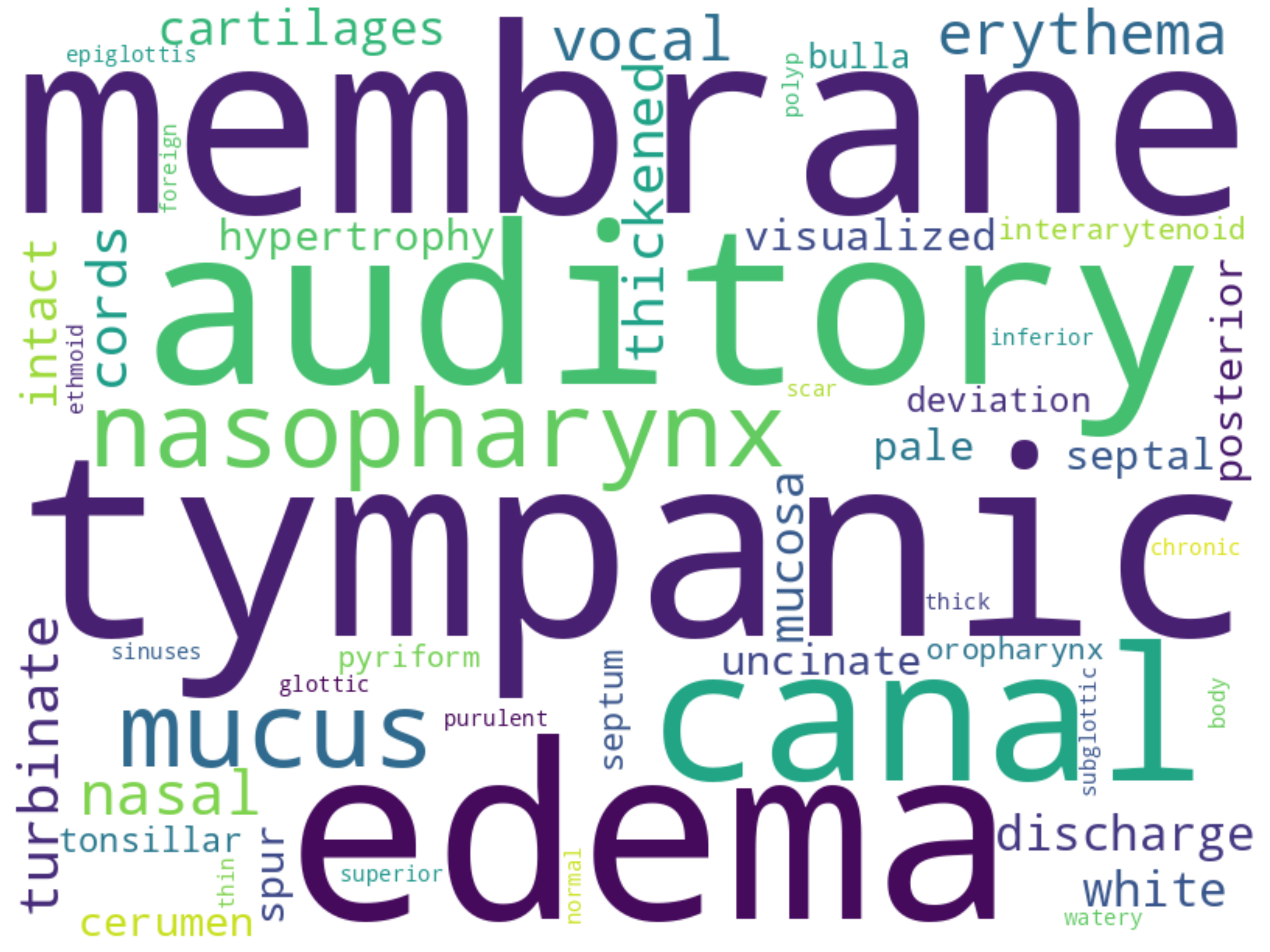}
    \caption{Word cloud visualization of text queries.}
    \label{fig:cloud}
\end{figure}
\subsection{ENTRep Dataset}
\noindent\textbf{Data Source.}
Images are acquired during clinical ENT endoscopy at \textit{Thong Nhat Hospital} (Ho Chi Minh City, Vietnam). These images reflect real-world practice across anatomical regions relevant to otolaryngology and include \texttt{normal} and \texttt{abnormal} findings.

\textbf{Data Curation.}
Clinical experts annotate each image for (i) anatomical region (\texttt{Classification}) and (ii) condition type (\texttt{Type} $\in\{\texttt{normal},\texttt{abnormal}\}$). When available, experts also provided a Vietnamese description with an English translation used for retrieval tasks. An excerpt of the JSON entry schema is shown below:
\begin{verbatim}
{
  "Path": "13135998_240329090725185831_121_image03.png",
  "Classification": "nose-left",
  "Type": "abnormal",
  "Description": "mào vách ngăn (T)",
  "DescriptionEN": "a spur on the septum"
}
\end{verbatim}
\noindent

Here, \texttt{Path} is the filename; \texttt{Classification} encodes the anatomical region (e.g., Ear Right/Left, Nose Right/Left, Throat, VC-open, VC-closed) and \texttt{Type} records normal/abnormal status. In addition, \texttt{Description} and \texttt{DescriptionEN} are the clinical narratives.

\noindent\textbf{Data Statistics.}
To enable rigorous evaluation and reduce the risk of overfitting, the dataset is partitioned into three subsets, as illustrated in Fig.~\ref{fig:cls}: a \emph{training} set (used for model development), a \emph{public test} set (for leaderboard evaluation), and a \emph{private test} set (used for final ranking). Acquired under routine clinical conditions, the images reflect realistic variations in illumination, viewpoint, motion blur, and device-specific characteristics. The classification data exhibit moderate class imbalance: the "nose-right" (25.2\%) and ``nose-left'' (22.5\%) classes are the most prevalent, whereas the ``throat'' class is the least represented (6.3\%); the remaining classes each account for approximately 11–13\%. The public and private test sets closely mirror the class distributions and sample sizes of the training set (645 and 646 images, respectively), indicating stratified sampling and enabling robust evaluation of model generalization.

For the image-to-image retrieval task, the training split contains 141 image pairs over 141 images, including 140 bidirectional links (99.3\%). Both the public and private test splits consist of 139 pairs over 139 unique images, with 136 bidirectional links (97.8\%) in each.

In the text-to-image retrieval task, each subset contains 71 pairs. In addition, the average textual query length is 5.42 words. A word cloud illustrating the distribution of query terms is shown in Fig.~\ref{fig:cloud}.

\begin{table}[!t]
\centering
\caption{Task 1 (Image Classification): Accuracy, Precision, Recall, and F1 score (\%) on the public and private test sets. }
\label{tab:task_1}
\resizebox{\columnwidth}{!}{%
\begin{tabular}{@{}lcccc@{}}
\toprule
\multicolumn{1}{c}{\textbf{Team}} & \textbf{Accuracy} & \textbf{Precision} & \textbf{Recall} & \textbf{F1}    \\ \midrule
\multicolumn{5}{c}{\textbf{Public Test}}                         \\
\midrule
\textbf{WAS}             & {95.66}    & {95.68}     & {95.66}  & {95.65} \\
\textbf{Soft Mind\_AIO}  & {94.88}    & {94.92}     & {94.88}  & {94.88} \\
\textbf{STG001}          & {94.57}    & {94.58}     & {94.57}  & {94.57} \\
\textbf{ZJU APRIL}       & 94.42    & 94.45     & 94.42  & 94.42 \\
\textbf{AIO-COFFEE}      & 93.80    & 93.91     & 93.80  & 93.81 \\
\midrule
\multicolumn{5}{c}{\textbf{Private Test}}                        \\
\midrule
\textbf{WAS}             & {95.82}    & {95.86}     & {95.82}  & {95.82} \\
\textbf{Soft Mind\_AIO}  & {95.20}    & {95.24}     & {95.20}  & {95.20} \\
\textbf{ZJU APRIL}       & {94.74}    & {94.80}     & {94.74}  & {94.74} \\
\textbf{AIO-COFFEE}      & 94.74    & 94.78     & 94.74  & 94.74 \\
\textbf{HCMUS-Shndrit!}  & 91.64    & 91.71     & 91.64  & 91.63 \\ \bottomrule
\end{tabular}%
}
\end{table}

\begin{table*}[!t]
\centering

\begin{minipage}[t]{0.49\textwidth}
\centering
\caption{Task 2 (Image-to-Image Retrieval): Recall@1 and Mean Reciprocal Rank (MRR, \%) on the public and private test sets.}
\label{tab:task_2}
\resizebox{\textwidth}{!}{%
\begin{tabular}{@{}lrr@{}}
\toprule
\multicolumn{1}{c}{\textbf{Team}} & \textbf{Recall@1} & \textbf{MRR} \\ 
\midrule
\multicolumn{3}{c}{\textbf{Public Test}} \\
\midrule
Soft Mind\_AIO   & {94.51} & {96.88} \\
Re:zero Slavery  & {93.38} & {96.65} \\
STG001           & {93.33} & {96.52} \\
NoResources      & 92.09 & 95.56 \\
ELO              & 91.82 & 95.25 \\
\midrule
\multicolumn{3}{c}{\textbf{Private Test}} \\
\midrule
Soft Mind\_AIO   & {92.09} & {95.70} \\
STG001           & {88.79} & {94.32} \\
HCMUS-Shndrit!   & {88.53} & {93.71} \\
Re:zero Slavery  & 88.42 & 94.11 \\
ELO              & 88.31 & 93.30 \\
\bottomrule
\end{tabular}%
}
\end{minipage}
\hfill
\begin{minipage}[t]{0.49\textwidth}
\centering
\caption{Task 3 (Text-to-Image Retrieval): Recall@1 and Mean Reciprocal Rank (MRR, \%) on the public and private test sets. }
\label{tab:task_3}
\resizebox{.89\textwidth}{!}{%
\begin{tabular}{@{}lrr@{}}
\toprule
\multicolumn{1}{c}{\textbf{Team}} & \textbf{Recall@1} & \textbf{MRR} \\ 
\midrule
\multicolumn{3}{c}{\textbf{Public Test}} \\
\midrule
H3N1           & {95.11} & {97.52} \\
ZJU APRIL      & 94.96 & 97.44 \\
SoloL          & 92.64 & 95.81 \\
NoResources    & 92.61 & 96.06 \\
STG001         & 92.25 & 95.65 \\
\midrule
\multicolumn{3}{c}{\textbf{Private Test}} \\
\midrule
SoloL          & {92.64} & {95.81} \\
ELO            & {90.77} & {95.12} \\
entropy        & {90.67} & {94.95} \\
STG001         & 89.79 & 94.81 \\
H3N1           & 85.56 & 91.58 \\
\bottomrule
\end{tabular}%
}
\end{minipage}

\end{table*}
\subsection{Evaluation Metrics}

\noindent\textbf{Task 1 (Image Classification).} We utilize \emph{Accuracy}, \emph{Precision}, \emph{Recall}, and \emph{F1-score}, computed with respect to the predicted anatomical region labels. In particular, per-class metrics are aggregated using a \emph{weighted average} based on class support, as defined in Eqn.~\eqref{eqn:cls}.
\begin{equation}
    \mathrm{Metric}_{\text{weighted}}
    = \sum_{c \in \mathcal{C}} \frac{n_c}{N}\,\mathrm{Metric}_c,
    \label{eqn:cls}
\end{equation}
where $n_c$ is the number of samples in class $c$, $\mathcal{C}$ is the set of classes, and $N=\sum_{c \in \mathcal{C}} n_c$ is the total number of samples. \emph{Accuracy} is reported as the overall fraction of correctly classified images.

\noindent\textbf{Task 2 (Image-to-Image Retrieval).}
We report \emph{Recall@1} and \emph{Mean Reciprocal Rank (MRR)}. A retrieved item is considered relevant if it matches the image. For a query $q$, let $r_q$ is the rank of its first relevant item (with $1/r_q=0$ if none is found), as defined in Eqn.~\eqref{eqn:mrr}.
\begin{equation}
    \mathrm{R@1} \;=\; \mathbb{1}[r_q=1], 
    \qquad
    \mathrm{MRR} \;=\; \frac{1}{N}\sum_{q=1}^{N}\frac{1}{r_q},
    \label{eqn:mrr}
\end{equation}
where $N$ is the number of queries and $\mathbb{1}[\cdot]$ is the indicator function.

\noindent\textbf{Task 3 (Text-to-Image Retrieval).} Following the evaluation protocol of Task 2, we report \emph{Recall@1} and \emph{Mean Reciprocal Rank (MRR)}, using the text query as input. An image is considered relevant if it constitutes a ground-truth match for the corresponding description.

\section{Evaluation Result}\label{sec:eval}

\noindent\textbf{Task 1: Image Classification.} As reported in Table~\ref{tab:task_1}, performance across both data splits is tightly clustered among the top-ranked models. \textit{WAS} achieves the highest scores on the public test set and maintains a slight lead on the private test set. \textit{Soft Mind\_AIO} consistently ranks second, followed closely by \textit{STG001} and \textit{ZJU APRIL}. The close alignment of Accuracy, Precision, Recall, and weighted F1 scores indicates balanced classification performance across classes. Moreover, the minimal performance degradation from the public to the private test set suggests strong generalization capabilities and limited susceptibility to leaderboard overfitting. In contrast, lower-ranked submissions (e.g., \textit{HCMUS-Shndrit!}) display more substantial declines, likely due to sensitivity to class imbalance and difficulties in resolving left/right distinctions or VC-state ambiguities.

\noindent\textbf{Task 2: Image-to-Image Retrieval.} As shown in Table~\ref{tab:task_2}, \textit{Soft Mind\_AIO} achieves the highest early precision across both data splits, while \textit{Re:zero Slavery} and \textit{STG001} demonstrate comparable performance on the public set. On the private set, model performance converges within a narrower range of 88–89\% in terms of Recall@1, indicating a tightly clustered outcome. The modest decline in performance from the public to private split suggests a limited distributional shift. Importantly, performance gains are concentrated at rank 1, underscoring the critical role of high-quality representations, symmetric pairing losses, and robust hard-negative mining strategies in this near one-to-one retrieval context.

\noindent\textbf{Task 3: Text-to-Image Retrieval.} As presented in Table~\ref{tab:task_3}, performance rankings shift between splits: H3N1 achieves the highest score on the public test set, while \textit{SoloL} ranks first on the private test. The larger generalization gap observed in some methods (e.g., \textit{H3N1}) suggests sensitivity to variations in textual description phrasing and coverage. In contrast, systems exhibiting more stable \texttt{MRR} across splits appear more robust to paraphrasing and synonym variation. These findings indicate that incorporating text normalization techniques and employing more powerful language encoders may enhance robustness and improve private-set performance stability.

\section{Administrative Details}\label{sec:detail}
\textbf{Organization.} The ENTRep challenge is organized by the \textit{Software Engineering Laboratory (SELab)}, University of Science, Viet Nam National University Ho Chi Minh City (VNU–HCM). The organizing team has extensive experience in medical imaging~\cite{tran2023support, dao2024vision, dao2024artificial, dao2024improving, phung2022disease, nguyen2023collaborative, le2024medgraph} and has recently explored applying large language models  and vision–language models to medical analysis~\cite{huynh2024dermai, tran2025enhancing}. In addition, the effort is conducted in collaboration with the University of Dayton (USA) and Thong Nhat Hospital (Ho Chi Minh City, Vietnam).

\noindent\textbf{Participation and data policy.} Use of the dataset requires agreement to the \textit{Dataset Usage Guideline}. Unless explicitly allowed, external private data are prohibited; publicly available pretrained weights may be used if documented. Organizers may audit submissions for compliance. Public leaderboards reflect public‑test performance; final rankings are based solely on private‑test scores.

\section{Conclusion and Outlook}\label{sec:conclusion}
ENTRep introduces a clinically benchmark for ENT endoscopy, combining fine-grained anatomical classification with both intra-modal and cross-modal retrieval tasks. Built on expert annotations and bilingual clinical narratives, it enables rigorous evaluation of models in a setting where accuracy and interpretability understanding are essential. A standardized output schema and server-side evaluation across development and test splits ensure reproducibility, while intuitive metrics align with real-world diagnostic needs. Initial results demonstrate that anatomy-aware designs help reduce common confusions, such as left–right laterality and vocal fold state errors. Furthermore, representational improvements primarily enhance early retrieval precision, which is especially impactful for clinical case lookup. Text-based systems benefit from synonym normalization and cross-lingual alignment, underscoring the value of structured label supervision paired with text descriptions.

Looking ahead, our ENTRep challenge highlights several directions for advancing vision–language models in clinical endoscopy. On the data front, expanding beyond a single institution, capturing device metadata, and incorporating short video clips could introduce domain variability and enable modeling of temporal dynamics. At the task level, adding hierarchical labels, defining soft relevance (e.g., via anatomical proximity), and reporting calibration and latency would better align with clinical decision-making and deployment constraints. Methodologically, improving generalization may involve domain-adaptive pretraining, anatomy-aware objectives, lightweight adaptation modules, and federated learning approaches. Enhanced bilingual querying via structured ontologies may reduce language mismatch, while clinician-in-the-loop evaluation can surface nuanced failure modes not captured by metrics.

\balance



\bibliographystyle{ACM-Reference-Format}
\bibliography{main}
\end{document}